\documentclass[acmtog]{acmart}
\usepackage{booktabs} 
\definecolor{my_href_pink}{RGB}{219, 48, 122}
\usepackage{cleveref}
\usepackage[ruled]{algorithm2e} 

\SetAlFnt{\small}
\SetAlCapFnt{\small}
\SetAlCapNameFnt{\small}
\SetAlCapHSkip{0pt}
\usepackage{overpic}
\acmJournal{TOG}

\begin{document}

\author{Xin Yu}
\affiliation{%
 \institution{The University of Hong Kong}
 \country{Hong Kong}
}
\email{yuxin27g@gmail.com}

\author{Ze Yuan}
\affiliation{%
 \institution{Beihang University}
 \country{China}
}
\email{yuanze1024@buaa.edu.cn}

\author{Yuan-Chen Guo}
\affiliation{%
 \institution{VAST}
 \country{China}
}
\email{imbennyguo@gmail.com}

\author{Ying-Tian Liu}
\affiliation{%
 \institution{Tsinghua University}
 \country{China}
}
\email{liuyingt23@mails.tsinghua.edu.cn}

\author{Jianhui Liu}
\affiliation{%
 \institution{The University of Hong Kong}
 \country{Hong Kong}
}
\email{jhliu0212@gmail.com}

\author{Yangguang Li}
\affiliation{%
 \institution{VAST}
 \country{China}
}
\email{liyangguang256@gmail.com}

\author{Yan-Pei Cao}
\affiliation{%
 \institution{VAST}
 \country{China}
}
\email{caoyanpei@gmail.com}

\author{Ding Liang}
\affiliation{%
 \institution{VAST}
 \country{China}
}
\email{liangding1990@163.com}

\author{Xiaojuan Qi}
\authornote{Corresponding author.}
\affiliation{%
 \institution{The University of Hong Kong}
 \country{Hong Kong}
}
\email{xjqi@eee.hku.hk}


\setcopyright{acmlicensed}
\acmJournal{TOG}
\acmYear{2024} \acmVolume{43} \acmNumber{6} \acmArticle{213} \acmMonth{12}\acmDOI{10.1145/3687909}


\newcommand{\modelname}{{TEXGen}}
\newcommand{\blockname}{\color{green}{"name"}}

\title{TEXGen: a Generative Diffusion Model for Mesh Textures}

\begin{teaserfigure}
    \centering
    \includegraphics[width=\textwidth]{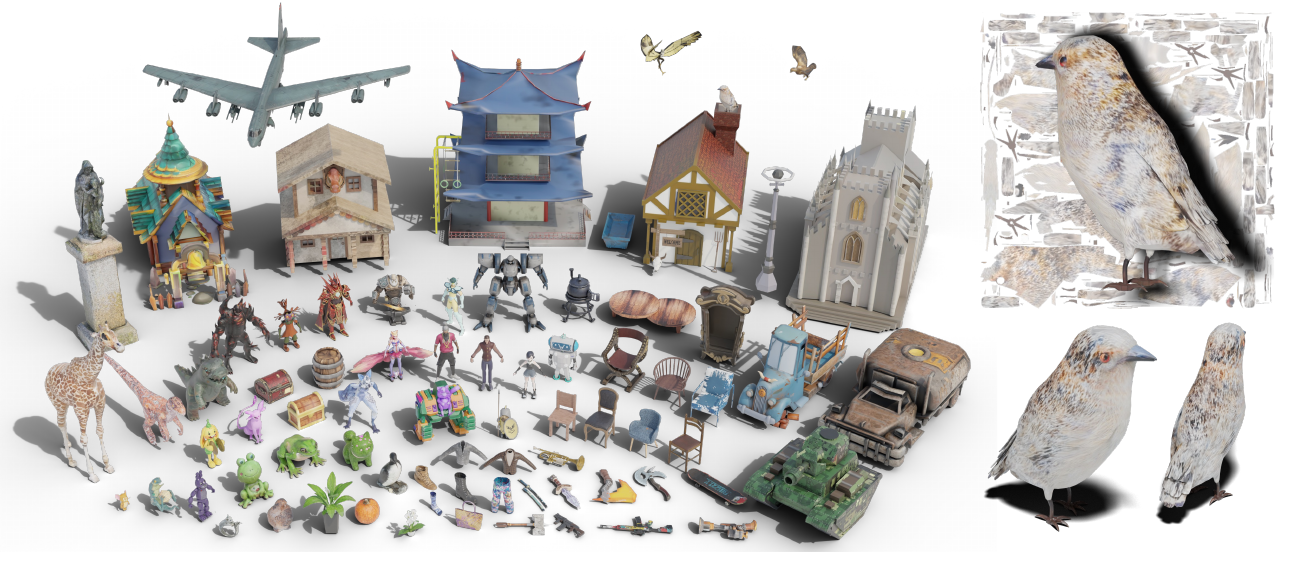}
    \caption{ 
    \textbf{3D meshes with textures generated by our method.}
    We show a gallery of 3D meshes with textures generated by our method (left) and the texture map and multi-view renderings of the bird model (right).
    Our approach models the distribution of mesh textures at high resolution, generating high-quality textures from text and image prompts, more multi-view renderings are shown in \cref{fig:indi}.
    }
\label{fig:teaser}
\end{teaserfigure}

\begin{abstract}
While high-quality texture maps are essential for realistic 3D asset rendering, few studies have explored learning directly in the texture space, especially on large-scale datasets. In this work, we depart from the conventional approach of relying on pre-trained 2D diffusion models for test-time optimization of 3D textures. Instead, we focus on the fundamental problem of learning in the UV texture space itself. For the first time, we train a large diffusion model capable of directly generating high-resolution texture maps in a feed-forward manner.
To facilitate efficient learning in high-resolution UV spaces, we propose a scalable network architecture that interleaves convolutions on UV maps with attention layers on point clouds. Leveraging this architectural design, we train a 700 million parameter diffusion model that can generate UV texture maps guided by text prompts and single-view images. Once trained, our model naturally supports various extended applications, including text-guided texture inpainting, sparse-view texture completion, and text-driven texture synthesis. Project page is at \textcolor{my_href_pink}{\href{https://cvmi-lab.github.io/TEXGen/}{https://cvmi-lab.github.io/TEXGen/}}.
\end{abstract}

\maketitle
\section{Introduction}
\label{sec:intr}

Synthesizing textures for 3D meshes is a fundamental problem in computer graphics and vision,  with numerous applications in virtual reality, game design, and animation. However,  the most advanced learning-based methods \cite{Yu_2023_ICCV,oechsle2019texture,siddiqui2022texturify,cheng2023tuvf} are restricted to generating textures for specific categories due to scalability and data limitation.
Recently, test-time optimization-based methods have emerged, which utilize pre-trained 2D diffusion models to produce image priors via score distillation sampling \cite{poole2022dreamfusion,yu2023text,lin2023magic3d,wang2023prolificdreamer} or by synthesizing pseudo-multi-views \cite{richardson2023texture,zeng2023paint3d,chen2023text2tex}. 
While these methods can generate textures for a wide range of objects,
they suffer from certain drawbacks, such as time-consuming per-object optimization and parameter tuning, susceptibility to the limitations of 2D priors, and poor 3D consistency in texture generation.

In recent years, there has been a surge in the development of large models across various domains, including natural language processing ~\cite{achiam2023gpt,touvron2023llama}, image and video generation ~\cite{betker2023improving,saharia2022photorealistic,blattmann2023stable}, and 3D creation \cite{hong2023lrm,li2023instant3d,xu2023dmv3d,wang2023pf,zou2023triplane,tochilkin2024triposr}. These models produce high-quality results and demonstrate remarkable generalization capabilities. Their success can be primarily attributed to two key factors: (1) scalable and effective network architectures that improve performance as model size and data amount increase, 
and (2) large-scale datasets that facilitate generalization.
In this paper, we explore the potential of building a large generative model by scaling up model size and data for generalizable and high-quality mesh texturing.

We introduce \modelname, a large generative model for mesh texturing. Our model utilizes a UV texture map as the representation for generation, as it is scalable and preserves high-resolution details.
More importantly, it enables direct supervision from ground-truth texture maps without solely relying on rendering loss \cite{hong2023lrm,li2023instant3d}, making it compatible with diffusion-based training and improving the overall generative quality. Previous works such as Point-UV-Diffusion \cite{Yu_2023_ICCV} and Paint3D \cite{zeng2023paint3d} have attempted to leverage diffusion models to learn the distribution of mesh textures. However, neither of these approaches achieved end-to-end training or feed-forward inference on general object datasets \cite{deitke2023objaverse}, resulting in error accumulation and scalability issues.

To perform effective feature interaction on mesh surfaces, we propose a scalable 2D-3D hybrid network architecture that incorporates convolution operations in the 2D UV space, followed by sparse convolutions and attention layers operating in the 3D space. This simple yet effective architecture offers several key advantages: (1) by applying convolution operations in the UV space, the network effectively learns local and high-resolution details; and (2) by further elevating the computation into 3D space, the network can learn global 3D dependencies and neighborhood relationships that are disrupted by the UV parameterization process, ensuring global 3D coherency. 
This hybrid design allows us to use sparse features in 3D space instead of dense voxel \cite{chen2019text2shape} or point features \cite{Yu_2023_ICCV,nichol2022point} for manageable computations while still maintaining 3D continuity, making the architecture scalable. 
By stacking multiple blocks, we train a large texture diffusion model capable of directly synthesizing high-resolution textures (e.g., 1024$\times1024$ texture maps) in a feed-forward manner guided by single-view images and text prompts. Moreover, our pre-trained model enables various applications, including text-guided texture synthesis, inpainting, and texture completion from sparse views.

To summarize, our contributions are as follows:
\begin{itemize}
    \item We introduce a novel network architecture designed for learning high-resolution UV texture maps, wherein we build a hybrid 2D-3D denoising block for effective feature learning. 
    \item Based on this architecture, we have trained a large diffusion model for high-resolution texture map generation. To the best of our knowledge, this is the first work capable of generating texture maps in an end-to-end manner without requiring additional stages, or test-time optimization.
    \item Our method achieves state-of-the-art results and serves as a foundation model supporting various training-free applications, such as text-guided texture synthesis, inpainting, and texture completion from sparse views.
\end{itemize}

\section{Related Work} 
\label{sec:rela}

\paragraph{Texture generation via 2D diffusion models.}
A prevalent method for texturing 3D meshes involves test-time optimization with pretrained 2D diffusion models. Techniques such as those based on score distillation sampling \cite{poole2022dreamfusion,latentnerf,fantasia3d, lin2023magic3d,wang2023prolificdreamer,yu2023text,yeh2024texturedreamer}, synthesize textures on 3D shapes by distilling 2D diffusion priors. However, these approaches have significant drawbacks, including high computational demands and inherent artifacts like the Janus problem and unnatural color. Another line of approach \cite{richardson2023texture,chen2023text2tex,cao2023texfusion,wu2024texro,gao2024genesistex,liu2023text,zhang2024mapa,ceylan2024matatlas} leverages geometry-conditioned image generation and inpainting to progressively generate the textures. TEXTure \cite{richardson2023texture}, for example, generates a partial texture map from one perspective view before using inpainting to complete other views. However, this method struggles with inconsistencies due to the lack of global information across views. Text2Tex \cite{chen2023text2tex} introduces an automated strategy for optimized viewpoint selection to avoid manual intervention. Meanwhile, TexFusion \cite{cao2023texfusion} proposes to aggregate appearances from multiple viewpoints during the diffusion denoising steps, producing a more consistent and cohesive texture map. Despite the advances, these methods predominantly lack 3D awareness due to their reliance on 2D diffusion models and often require time-consuming per-instance optimization.

\paragraph{Texture generative models.}
A variety of learning-based approaches have been developed to train generative models from 3D data \cite{chang2015shapenet} for mesh texturing. Early methods \cite{oechsle2019texture} introduced implicit texture fields that assign colors to each point on the 3D surface. However, these methods often struggle to reproduce high-frequency details due to the continuous nature of implicit fields. Texturify~\cite{siddiqui2022texturify} and Mesh2Tex~\cite{bokhovkin2023mesh2tex} design convolution operations tailored for mesh structures to facilitate learning directly on surfaces, which use the StyleGAN architecture \cite{karras2019style} to predict textures for each mesh face and relies on GANs \cite{goodfellow2020generative} for training. Despite these advances, these methods are susceptible to mode collapse due to the instability of GAN training. More recent approaches, such as TUVF \cite{cheng2023tuvf} and Point-UV Diffusion \cite{Yu_2023_ICCV}, attempt to generate UV maps directly for 3D shapes, addressing some of the aforementioned challenges. However, these methods are generally limited to category-specific objects and struggle with generalized objects. Paint3D has demonstrated the capability of handling generalized objects by fine-tuning a diffusion model for texture maps on larger-scale datasets \cite{deitke2023objaverse}. Nonetheless, it still requires test-time optimization to generate the initial textures, and the trained diffusion model is only capable of removing light effects and filling holes. The two-stage pipeline can lead to cumulative quality losses, often resulting in degenerated details in the final output.

\paragraph{Feed-forward methods for 3D generation.}
Recently, there has been a notable shift in the community towards training feed-forward 3D generative models using large-scale datasets. These models are designed to accept minimal input conditions and directly output 3D representations, thereby eliminating the need for per-instance optimization \cite{poole2022dreamfusion,sjc,latentnerf,fantasia3d, lin2023magic3d,wang2023prolificdreamer,yu2023text}. Notably, Large Reconstruction Model (LRM) and its variants \cite{hong2023lrm, li2023instant3d, grm, zou2023triplane, tochilkin2024triposr} adopt a transformer-based architecture to infer 3D shapes from single or sparse-view inputs, which show significant improvements on the quality and efficiency of feed-forward 3D reconstruction. However, these methods often result in over-smoothed appearances, especially in areas not visible in the input views, and lack the capability to produce varied outcomes by design. Furthermore, adapting these models for texture generation given the 3D geometry poses significant challenges, as they typically manage feature interactions in a coarse-grained 3D space rather than directly on surfaces.

\section{Overview}
Given a 3D mesh $S$, our objective is to develop a generative model capable of producing high-quality textures for 3D surfaces based on user-defined conditions such as images or text prompts, as illustrated in \cref{fig:pipeline} (a). The modeling comprises the following principal steps:

\textbf{(i) Data representations.} 
We use UV texture maps as the mesh texture representation, which is compact and suitable for diffusion training. We discuss its characteristics in \cref{subsec:representation} which motivate us to develop a new network architecture in \cref{subsec:network}. 

\textbf{(ii) Model construction and learning.} 
We develop a novel hybrid 2D-3D network structure that effectively handles the unique characteristics of texture maps (\cref{subsec:network}). We then train a diffusion model~\cite{ho2020denoising} to generate high-resolution texture maps for a given mesh based on a single-view image and a text description (\cref{subsec:diffusion}). 

\textbf{(iii) Inference.}
After training is done, our model can start from a noise image and iteratively denoise it to generate high-resolution texture maps. Additionally, our model supports various training-free extensions, such as text-guided texture synthesis, texture inpainting, and texture completion from sparse views (\cref{subsec:generation}).

\section{Method}


\subsection{Representation for Texture Synthesis} 
\label{subsec:representation}
\begin{figure}[ht]
    \includegraphics[trim={0cm 0cm 0cm 0cm},clip,width=0.9\columnwidth]{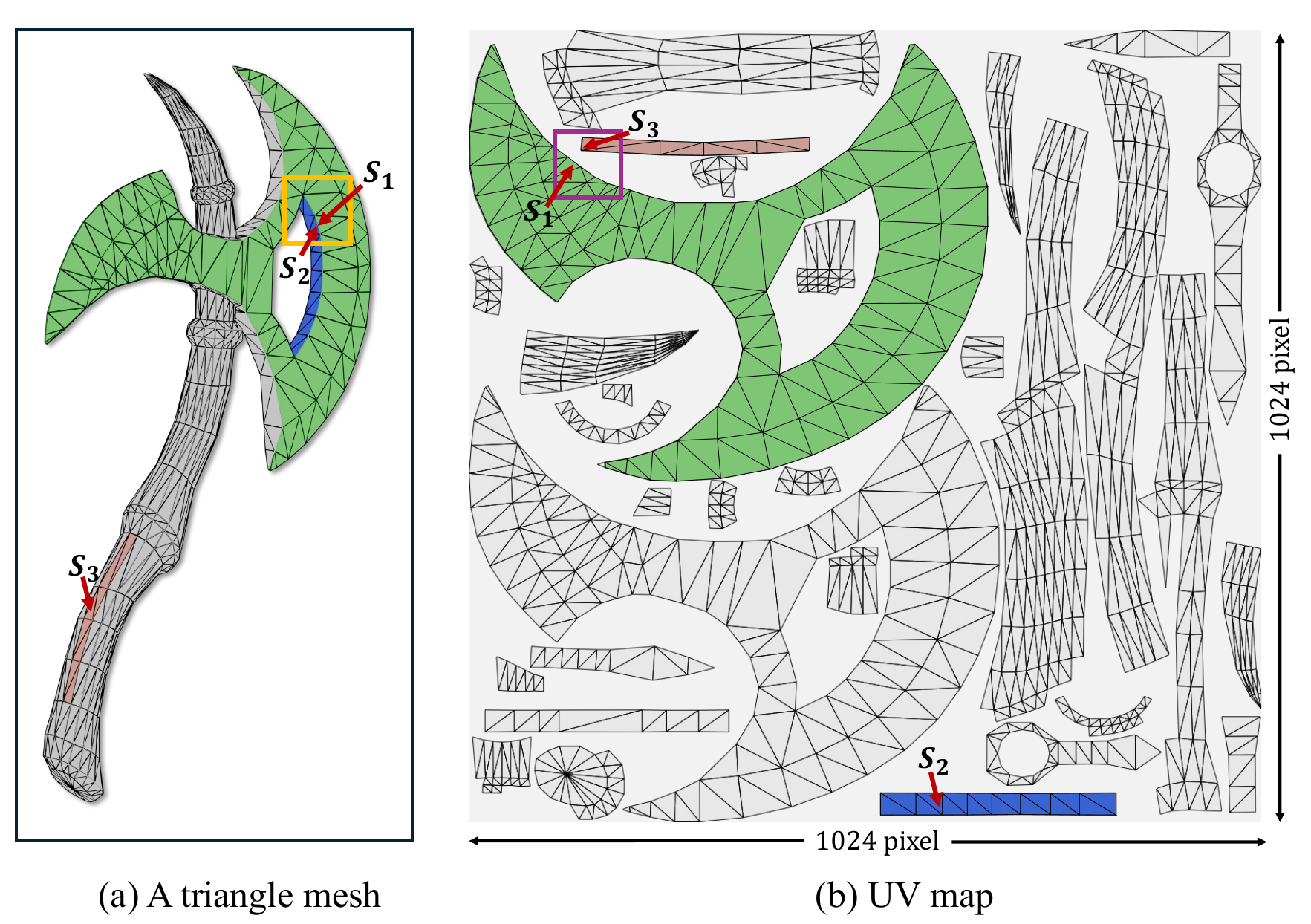}
    \vspace{-2mm}
    \caption{
    \textbf{An illustration of (a) a mesh with its (b) UV map.} Three islands $S_1$, $S_2$ and $S_3$ are shown both on the mesh surface and its flattened UV map, where continuous islands $S_1$ and $S_2$ are positioned far apart on the UV map while disconnected islands $S_1$ and $S_3$ show closer distance on the UV map.}
    \label{fig:uv}
    \vspace{-4mm}
\end{figure}

A surface can be fundamentally viewed as a two-dimensional signal embedded within a three-dimensional space. Consequently, a traditional technique in graphics for processing mesh structures is UV mapping, which flattens the 3D structure into a compact 2D representation (refer to \cref{fig:uv}). This transformation allows 3D attributes, such as textures, to be reorganized and represented on a 2D plane. The 2D UV space effectively captures neighborhood dependencies within individual islands, enhancing computational efficiency for texture generation \cite{Yu_2023_ICCV} thanks to its grid structure. Additionally, the explicit nature of the texture map facilitates direct supervision, making it well-suited for integration with diffusion models.

The advantages outlined above motivate our adoption of a 2D UV texture map as the representation for texturing 3D meshes. However, despite its merits, this approach inevitably loses the global-level 3D consistency among different islands due to the fragmentation inherent in UV mapping. As illustrated in \cref{fig:uv}, islands \(S_1\) and \(S_2\) are contiguous on the 3D surface but are positioned far apart on the UV map. Conversely, \(S_1\) and \(S_3\), which are adjacent on the UV map, do not share a physical connection on the surface. This fragmentation can lead to inaccurate feature extraction in conventional image-based models. To address this issue, we propose a novel model that synergizes the strengths of the 2D UV space—enabling high-resolution and detailed feature learning—with the incorporation of 3D points to maintain global consistency and continuities. These components interleave and refine representations, facilitating effective learning for generating high-resolution 2D texture maps. Further details will be in \cref{sec:model}.

\begin{figure*}[t]
    \includegraphics[trim={0cm 0cm 0cm 0cm},clip,width=\textwidth]{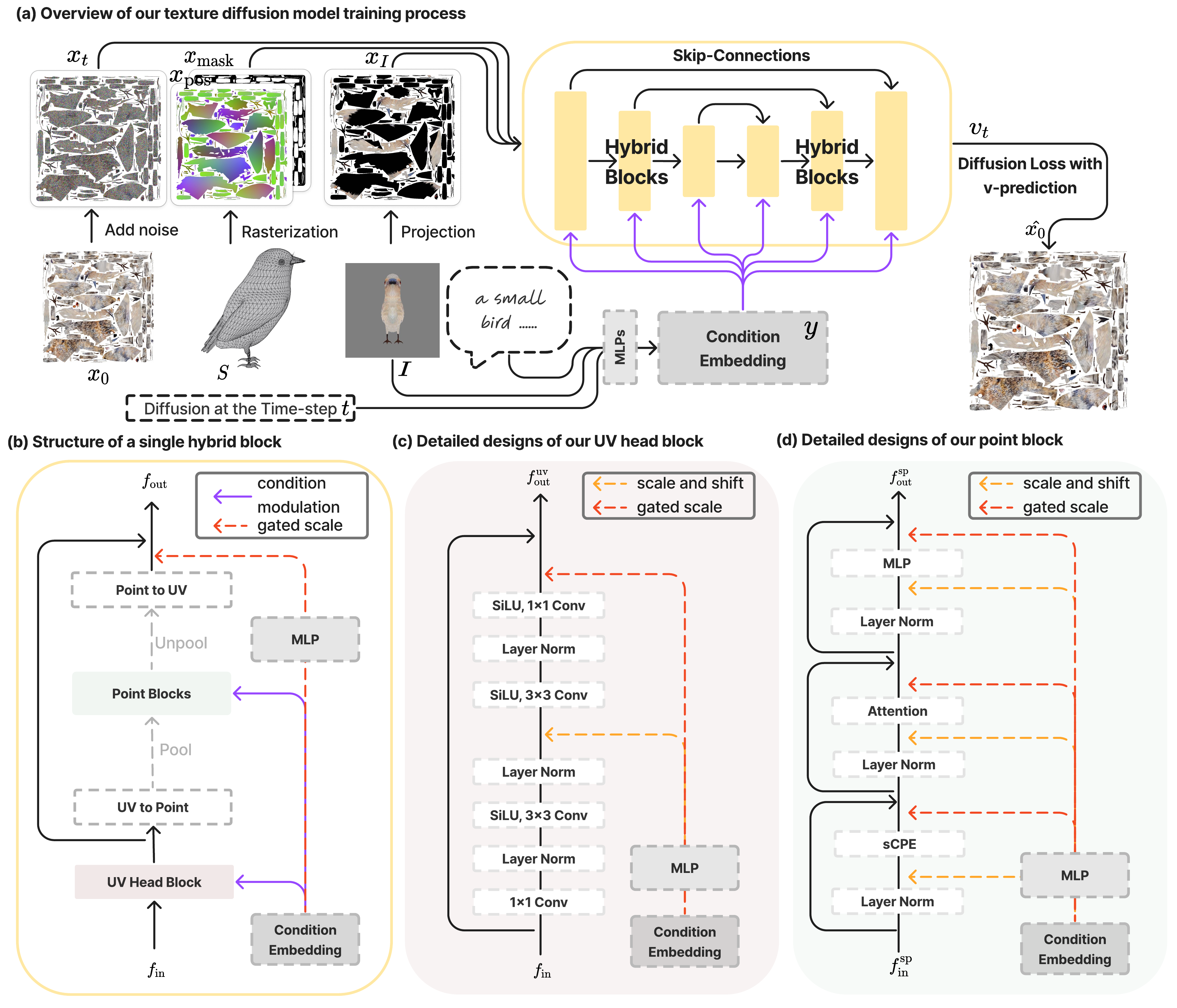}
    \caption{\textbf{An overview of TEXGen.} (a). An overview of our training pipeline. We train a diffusion model to
    generate high-resolution texture maps for a given mesh $S$ based on a single-view image $I$ and text descriptions by learning to denoise from a noise texture map $x_t$. The core of our denoising network is our proposed hybrid 2D-3D block. (b). The structure of a single hybrid block. (c)-(d). The detailed designs of our UV head block and point block.}
    \label{fig:pipeline}
\end{figure*}

\begin{figure}[t]
    \includegraphics[trim={0cm 0cm 0cm 0cm},clip,width=\columnwidth]{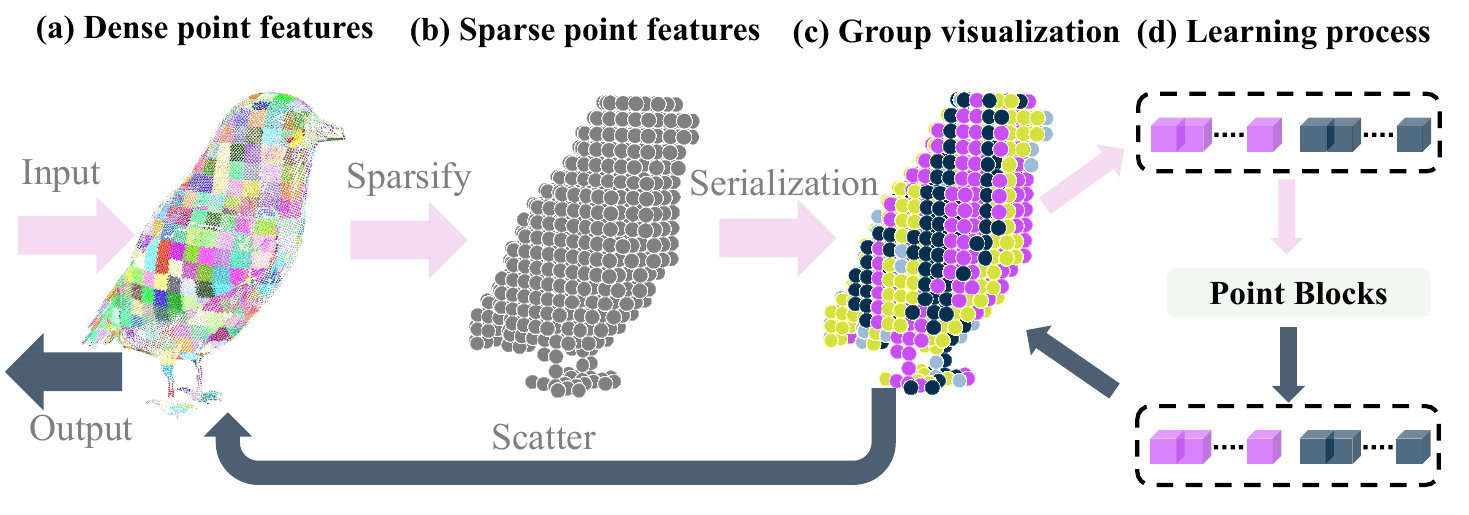}
    \caption{\textbf{An illustration of the feature learning procedure in 3D space.} In panel (a), we start with rasterized dense point features, which we sparsify using grid-pooling to create sparse point features shown in (b). Different pools are indicated by various colors in (a). These points are then serialized to determine their order for subsequent group-based self-attention, as part of the learning process shown in (d). In (c), we visualize different groups formed based on Hilbert serialization, where each color signifies a distinct group. Finally, the processed features are scattered back to their original coordinates, providing the output dense point features.}
    \label{fig:3d_process}
    \vspace{-4mm}
\end{figure}

\subsection{Model Construction}
\label{subsec:network}
 Utilizing 2D texture representations, we can train a diffusion model that conducts iterative denoising to generate a high-quality 2D texture map, given a specific condition, such as a posed single image or a text prompt. 
The core of our model is a hybrid 2D-3D network that learns features in both 2D and 3D spaces (see \cref{fig:pipeline}). 
Unlike unconditional generation, our work prioritizes conditional generation, particularly conditioning on textual and visual inputs. Text prompts provide an intuitive interface for users to specify desired attributes in the generated content, making the model more accessible and responsive to user intentions. On the other hand, conditioning on images offers precise control over the generation process by capturing pixel-level details that text alone may overlook, thus offering stronger guidance for diffusion models. Moreover, a single image with rich textures can serve as a valuable prior in the diffusion process, facilitating more effective learning. Since it is feasible to generate an image from text alone using text-to-image models~\cite{rombach2022high, zhang2023adding}, we choose to condition on both text and images for training. During inference, the model remains flexible, allowing for the inclusion or omission of image data based on its availability to the user (see \cref{subsec:generation}).

\label{sec:model}
\paragraph{Network}  
As depicted in \cref{fig:pipeline}, our training pipeline utilizes a diffusion-based approach. At each denoising step, our network processes multiple inputs: the noised texture map $x_t$, position map $x_{\text{pos}}$, mask map $x_{\text{mask}}$, a single image $I$, text prompt $c$, and timestep $t$, to guide the removal of noise from $x_t$. 
The integration of the image $I$ into the network occurs in two distinct ways: \textbf{(1) Projection of image pixels:} The image pixels are projected back onto the surface to derive a partial texture map $x_I$, which serves as an additional input. \textbf{(2) Global embeddings extraction:} Using an image encoder from CLIP \cite{radford2021learning} and a text encoder, we extract global image and text embeddings, respectively. A learnable timestep embedding accommodates different values of $t$. These embeddings are processed through separate MLPs and subsequently combined to form the global condition embedding $y$. This embedding modulates features within the network to incorporate condition-specific information, similar to \cite{peebles2023scalable}.
The network predicts the velocity $v_t$ \cite{salimans2022progressive}, which can be equivalently transformed into a prediction of noise $\epsilon$ or a prediction of $x_0$.
Similar to the denoising network described in \cite{ho2020denoising}, our architecture is based on the UNet framework \cite{ronneberger2015u}. However, we uniquely enhance it by incorporating hybrid 2D-3D blocks at each stage. This adaptation enables our network to adeptly manage the distinct characteristics of texture maps.

\paragraph{Hybrid 2D-3D block} 
The key to our design is the hybrid 2D-3D block, which facilitates efficient feature learning for 2D texture map generation. As shown in \cref{fig:pipeline} (b), our hybrid block comprises a UV head and several 3D point-cloud blocks. An input UV feature $f_{\text{in}}$ is first processed through a 2D convolution block (see \cref{fig:pipeline} (c)) to extract local features in the UV space. 2D convolutions are computationally more efficient compared to 3D convolutions or point cloud KNN searches for establishing neighbors and weighting, making it more scalable to higher resolution. Furthermore, within an island, 2D convolutions ensure that the aggregation of adjacent features is based on surface neighborhoods rather than volumetric neighborhoods, where geodesic distances can be larger.
Thus, this step efficiently ensures the preservation of high-resolution information.

To establish 3D connections among islands in UV space, we employ rasterization to remap the output UV features 
 $f^{\text{uv}}_{\text{out}}$ back into the 3D space, thus reorganizing these UV features into 3D point cloud features $f^{\text{point}}_{\text{in}}$. The primary objective in the 3D space is to acquire 3D neighborhood relationships and global structural features to improve 3D consistency rather than extracting high-resolution detail features. Consequently, we employ relatively sparse features and design an efficient module to ensure scalability. A brief illustration is shown in \cref{fig:3d_process}. The key components are detailed as follows:
\begin{itemize}    
    \item \textbf{Serialized attention.} For the input dense point features $f^{\text{point}}_{\text{in}}$, we adopt grid-pooling \cite{wu2022point} to sparsify the number of points and obtain $f^{\text{sp}}_{\text{in}}$. The pooled features are then treated as tokens and processed by point attention layers for learning. To boost the efficiency, we utilize Serialized Attention \cite{wu2023point}, which facilitates efficient patch-based attention. Specifically, the point features are partitioned into different groups based on their sterilized codes, defined by space-filling curves, such as the z-order curve \cite{morton1966computer} and the Hilbert curve \cite{hilbert1935neubegrundung}.

    \item \textbf{Position encoding.} 
    Position encoding plays a critical role in incorporating 3D position information into our model. Traditional methods that use point coordinates as cues for position encoding \cite{lai2022stratified,yang2023swin3d} are less effective compared to conditional positional encoding \cite{chu2021conditional,wang2023octformer,wu2023point}, which utilizes convolution layers for this purpose \cite{wu2023point}. Initially, we implemented the xCPE \cite{wu2023point}, which integrates a sparse convolution layer directly before the attention layer. However, this approach proved to be inefficient and time-consuming as the transformer dimension increases (i.e., \(d=2048\)). To address these inefficiencies, we developed a modified approach, termed sCPE, which uses a linear layer to reduce the input's channel dimension before executing the sparse convolution. Subsequently, another linear layer is used to expand the channel dimension back to its original size to match the feature dimension of the skip connection.

    \item \textbf{Condition modulation.} For both our 2D and 3D blocks, we utilize the global condition embedding to modulate intermediate features, enabling the injection of conditional information. Specifically, inspired by DiT \cite{peebles2023scalable}, we employ MLPs to learn modulation vectors $\gamma$ and $\beta$ from the condition embedding $y$. These vectors are used to scale and shift the intermediate features across their channel dimensions, formulated as $f_{\text{mod}} = (1 + \gamma)\cdot f_{\text{in}} + \beta$. Besides, we also learn a gated scale $\alpha$ to scale the output feature before its fusion with the feature from the skip connection, expressed as $f_{\text{fuse}} = \alpha\cdot f_{\text{out}} + f_{\text{skip}}$.

\end{itemize}

The learned sparse point features $f^{\text{sp}}_{\text{out}}$ are then scattered to dense coordinates based on grid partition, resulting in $f^{\text{point}}_{\text{out}}$. Before fusion with the skip-connected UV feature, we also learn a gated scale $\alpha^{\text{point}}$ from the condition embedding $y$ to scale the point features. The final fused feature is given by:
$ f_{\text{out}} = f^{\text{uv}}_{\text{out}} + \alpha^{\text{point}} \cdot f^{\text{point}}_{\text{out}} $

\subsection{Diffusion Learning}
\label{subsec:diffusion}
Given a real texture map $x_0$, we randomly sample a time-step $t$ ($t \in \{0, 1, \ldots, 1000\}$) and add noise to the texture map by
\begin{equation}
x_t = \sqrt{\bar{\alpha}_t} x_0 + \sqrt{1 - \bar{\alpha}_t} \epsilon,
\end{equation}
where $\epsilon \sim \mathcal{N}(0, \mathbf{I})$, and $\{\bar{\alpha}_t\}$ are hyperparameters that follow a certain noise scheduler. Specifically, we use a noise scheduler from Stable Diffusion \cite{rombach2022high} and adopt zero-terminal SNR \cite{lin2024common} to scale the original noise scheduler such that $\bar{\alpha}_t = 0$ when $t = 1000$. This helps eliminate the gap between training and inference at the initial starting point. During training, we randomly drop text embeddings and image embeddings with a probability $p=0.2$ so that we can utilize classifier-free guidance \cite{ho2022classifier} during inference.
For the network output $x_{\text{out}}$, we use v-prediction \cite{salimans2022progressive,lin2024common} to compute the diffusion loss, i.e.,
\begin{equation}
\label{eq:v-pred}
\begin{aligned}
    v_t &= \sqrt{\bar{\alpha}_t} \epsilon - \sqrt{1 - \bar{\alpha}_t} x_0, \\
    \mathcal{L}_{\text{diff}} &= \lambda_t \left\| v_t - x_{\text{out}} \right\|_2^2,
\end{aligned}
\end{equation}
where $\lambda_t$ is the soft-min-SNR weight~\cite{crowson2024scalable}.

Notably, we obtain the predicted sample $\hat{x_0}$ from the v-prediction output and apply an LPIPS~\cite{zhang2018unreasonable} loss on multi-view renderings for extra supervision:
\begin{equation}
\mathcal{L}_{\text{render}} = \frac{1}{N} \sum \text{LPIPS}(\hat{I_i}, I_i),
\end{equation}
where $\hat{I_i}$ is an image rendered from a random viewpoint using the predicted texture map $\hat{x_0}$, and $I_i$ is the corresponding ground truth image. Our final loss is:

\begin{equation}
\mathcal{L} = \lambda_1 \mathcal{L}_{\text{diff}} + \lambda_2 \mathcal{L}_{\text{render}},
\end{equation}
where we set $\lambda_1 = 1$ and $\lambda_2 = 0.5$.

\subsection{Texture Generation}
\label{subsec:generation}

\begin{figure*}
    \includegraphics[trim={0cm 0cm 0cm 0cm},clip,width=\textwidth]{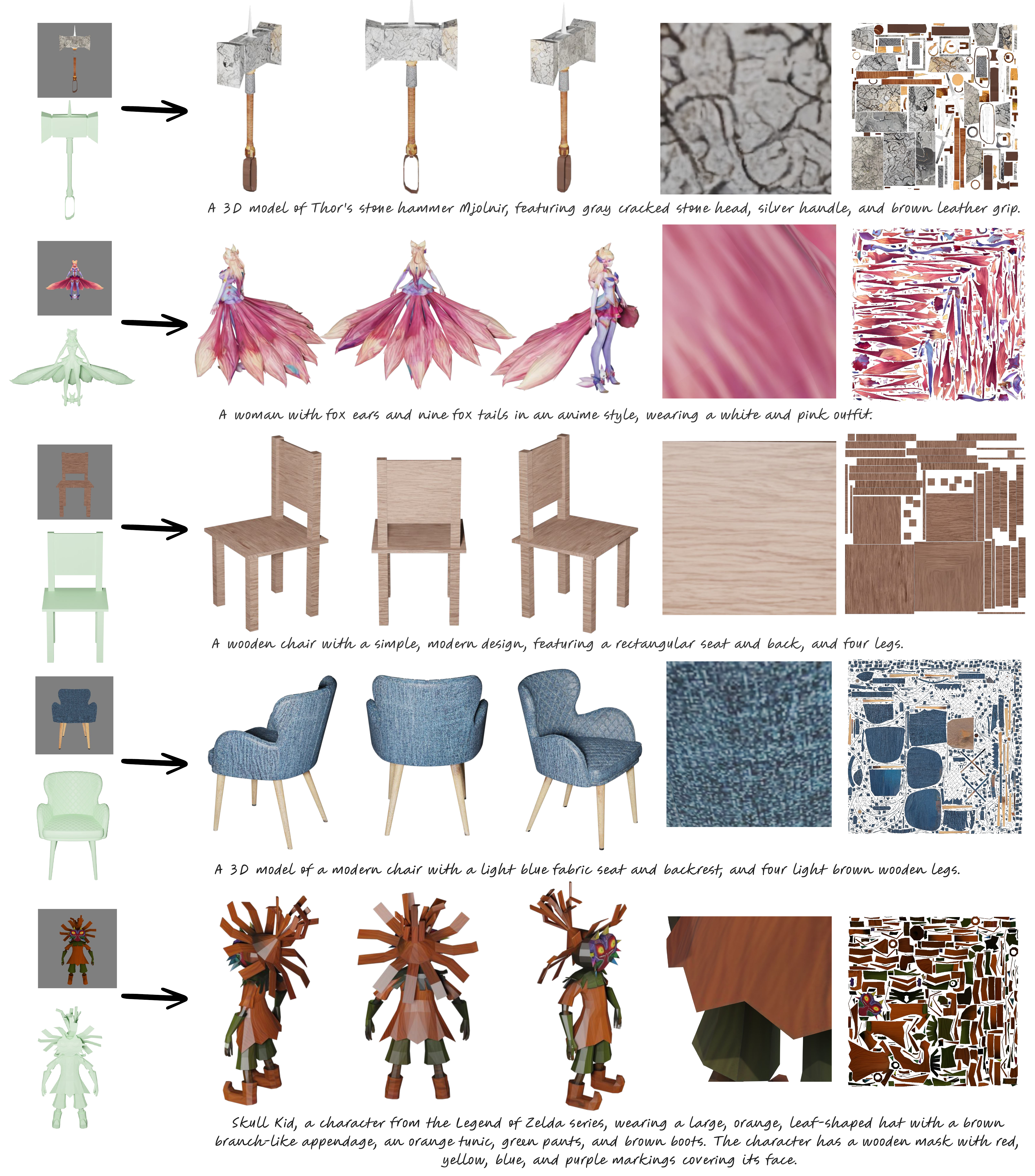}
    \caption{\textbf{Texture generation results.} For given meshes, our method can synthesize highly detailed textures conditioned on guided single-view images and text prompts. We show three novel view images from our textured results and representative zoom-in regions from the textured mesh. The generated full texture maps are also shown.}
    \label{fig:indi}
\end{figure*}

After training, our denoising network is ready to generate high-quality texture maps for 3D meshes (see \cref{fig:teaser} and \cref{fig:indi}). We start by initializing a pure Gaussian noise texture map in UV space. Then, using conditional information (e.g., single-view image, text prompt), we iteratively denoise it to generate a final texture map. To accelerate inference, we use DDIM \cite{song2020denoising} sampling for 30 steps. Interestingly, although we trained our model guided by a single-view image and text prompt, it can be generalized to other scenarios and applications during testing.

\paragraph{Text to texture generation.} If only a text prompt is provided, we can arbitrarily choose a mesh viewpoint, render a depth map, and then use ControlNet \cite{zhang2023adding} to generate a corresponding single-view image. This is also a motivation for us to train an image-conditioned model instead of a text-conditioned model since an image can be easily obtained from text, whereas a text-conditioned model lacks the control capabilities provided by an image.

\paragraph{Texture inpainting.}
During training, the pixel information of the single-view image is projected back to UV space, resulting in a partial initial texture map. Our network is trained to fill in the unseen parts. We found that this capability allows the model to function as a texture inpainting model during testing. Specifically, we can take the user-provided partial texture map and mask as $x_I$ (i.e., skipping the projection from the single-view step) and input them into the network for inpainting. For the image embedding required during testing, we set it to zero embedding, as our training included randomly dropping the image embedding, making the model robust to this situation.

\paragraph{Texture completion from sparse views.}
If the user provides a few sparse-view images, such as two images, we can effectively utilize the additional information for a generation. We simply project and fuse each image during the projection step and randomly select one image for image embedding extraction. Our model can fill textures in the occluded parts and recover the whole texture map.

\begin{figure*}[t]
    \includegraphics[trim={0cm 0cm 0cm 0cm},clip,width=\textwidth]{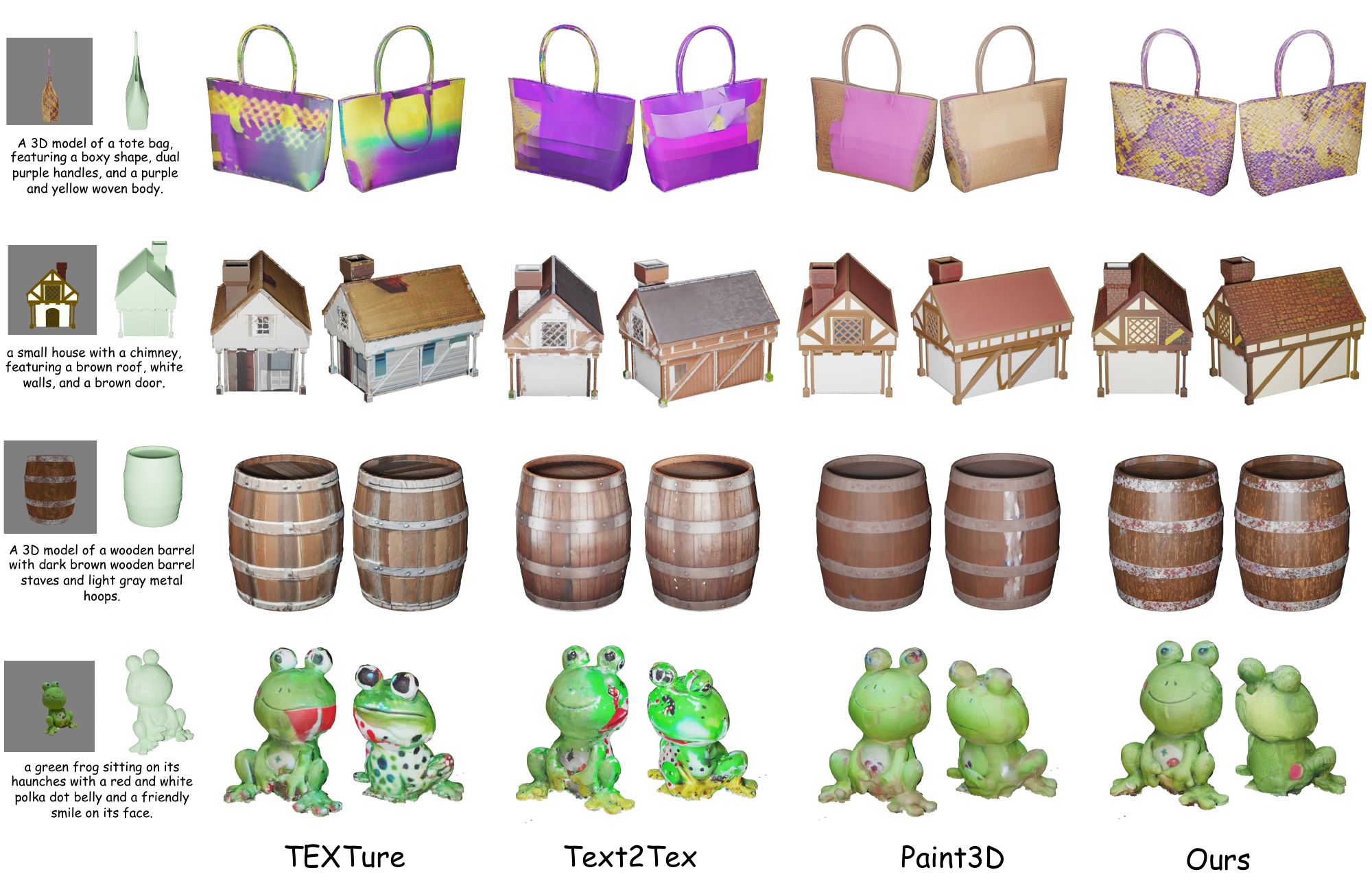}
    \caption{\textbf{Comparison with state-of-the-art methods.} We compare our method with four representative state-of-the-art methods. Our model can synthesize more detailed and coherent textures compared to these methods which rely on test-time optimization using a 2D pretrained text-to-image diffusion model. Also, our method trained on the 3D dataset and 3D representation avoids the Janus problem that commonly occurs in other methods.}
    \label{fig:comp}
\end{figure*}
\begin{figure*}[t]
    \includegraphics[trim={0cm 0cm 0cm 0cm},clip,width=\textwidth]{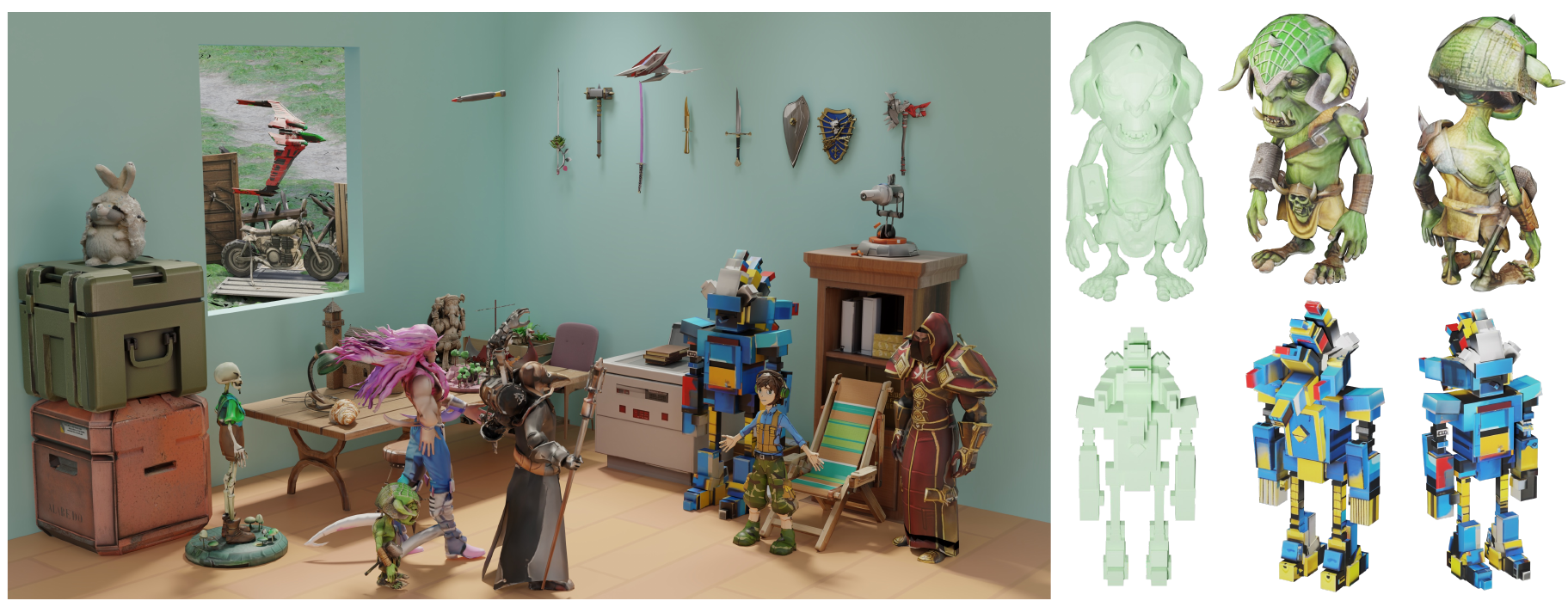}
    \caption{\textbf{An indoor scene with all meshes textured by \modelname.} We generate a single view using text-conditioned ControlNet with depth control for each mesh and paint them with both the text and single view prompt with \modelname.}
    \label{fig:text_condition}
\end{figure*}
\section{Experiments} 
\label{sec:expe}
We use Objaverse \cite{deitke2023objaverse} as our raw data source, which comprises over 800,000 3D meshes. After processing and cleaning this dataset, we extracted a total of 120,400 data pairs. Of these, 120,000 pairs are designated for training and the remaining 400 pairs are set aside for evaluation. We provide a detailed data processing method and implementation details of our model in the Appendix.

\subsection{Main Results and Comparisons}
We present our primary results, which include the textured 3D mesh conditioned on a single-view image and a text prompt, as illustrated in \cref{fig:teaser}. Notably, consider the example of the bird, where the texture detailing on the feathers demonstrates the model's capability to generate highly detailed textures. In \cref{fig:indi}, we display the conditions and multi-view results of several examples independently. Our results demonstrate that the model can generate high-quality textures with rich local details, preserve condition information, and achieve global coherence. We compare our method with other generalizable texture generation methods, including TEXTure \cite{richardson2023texture}, Text2Tex \cite{chen2023text2tex}, and Paint3D \cite{zeng2023paint3d}.

\paragraph{Qualitative comparisons.}
We conducted a qualitative analysis comparing our method with TEXTure and Text2Tex. Both these methods utilize a 2D pretrained text-to-image diffusion model for test-time optimization to texture 3D meshes. Their approach involves generating an image with a depth-conditioned diffusion model that aligns with the geometry of the current view, which is then projected onto the mesh. For areas without direct visibility, they iteratively adjust the viewpoint and employ an inpainting model to complete the texture. This cycle is repeated to produce a full texture map. For a fair comparison, we replaced their initial image with the same single-view image used in our method. As depicted in \cref{fig:comp}, both TEXTure and Text2Tex encounter several challenges. In the first example, despite the conditional image containing rich texture patterns, both methods produce overly smooth textures. In the second and third examples, while more details are shown, various artifacts degrade their quality. Besides, they do not well preserve the guided image information. In the fourth example, they face the Janus problem, where inappropriate features such as eyes and a mouth appear on both the front and back of the frog. In contrast, our method successfully generates textures with rich detail, maintains global coherence, and avoids the Janus problem.

Then, we compare our method with Paint3D, which utilizes a two-stage approach. In the first stage, similar to TEXTure and Text2Tex, Paint3D generates a texture map $x^{\text{coarse}}$ through iterative inpainting; additionally, it trains a refinement model $D$ to address unrealistic lighting issues and fill in areas that were not painted in the first stage. However, this model cannot be used standalone for texture generation and we observe an inevitable loss of texture details after the refinement stage. The results in \cref{fig:comp} show Paint3D produces over-smooth results and still exhibits the Janus problem.

\begin{table}[t]
\centering
\caption{\textbf{Quantitative Comparisons.} FID and KID ($\times 10^{-4}$) are evaluated on multi-view renderings. Our method achieves state-of-the-art texture quality with significantly faster inference.}
\begin{tabular}{cllll}
\toprule
Methods & FID($\downarrow$) & KID($\downarrow$) & Time($\downarrow$) \\
\midrule
TEXTure \cite{richardson2023texture} & 48.31 & 48.00 & 80s \\
Text2Tex \cite{chen2023text2tex} & 49.85 & 47.38 & 344s \\
Paint3D \cite{zeng2023paint3d} & 43.55 & 25.73 & 95s \\
Ours & \bf{34.53} & \bf{11.94} & \bf{10s} \\
\bottomrule
\end{tabular}
\vspace{-2mm}
\label{table:quan}
\end{table}

\begin{figure}[t]
    \includegraphics[trim={0cm 0cm 0cm 0cm},clip,width=\columnwidth]{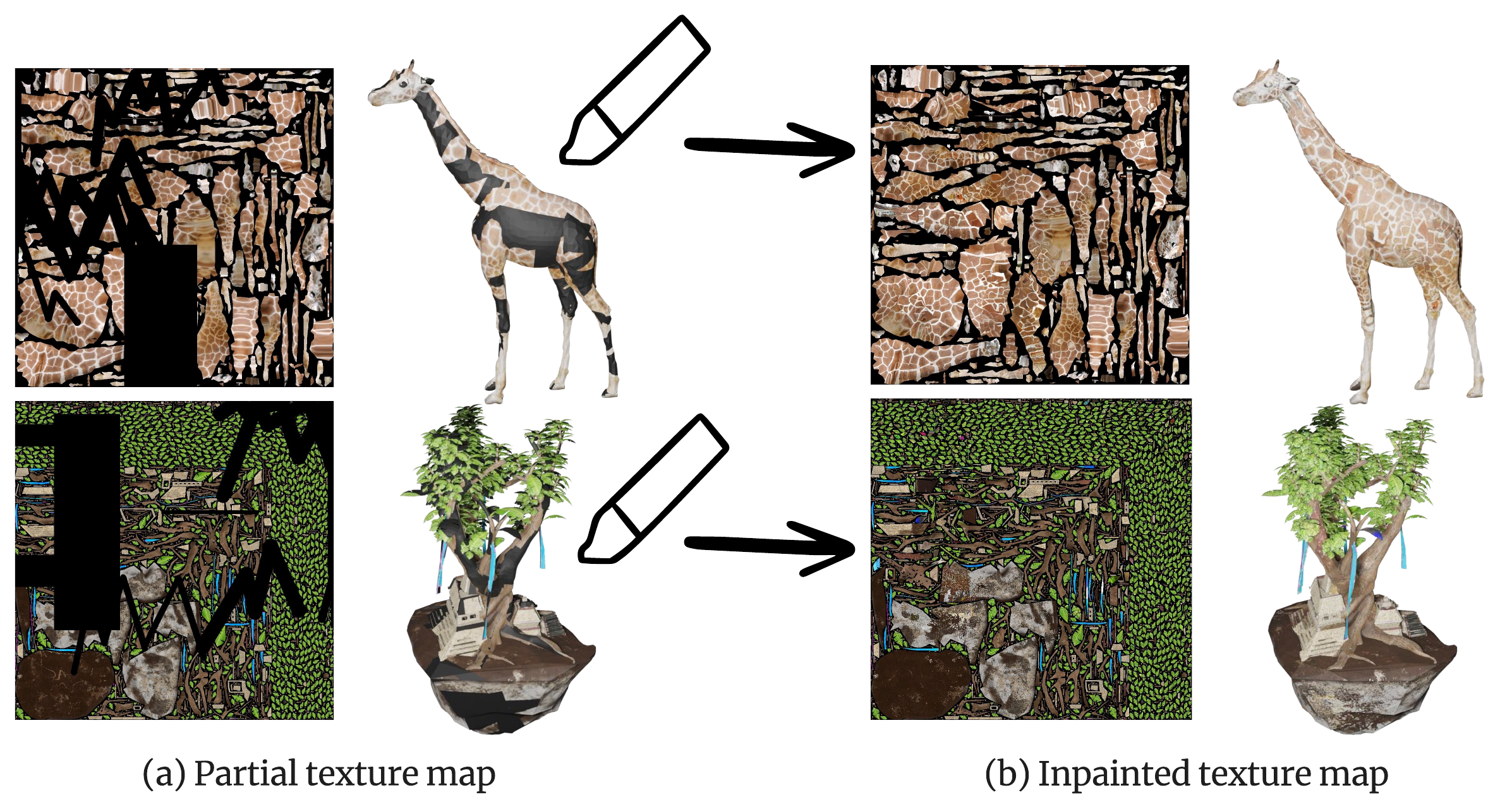}
    \caption{\textbf{TEXGen as a texture inpainter.} We demonstrate the potential of TEXGen as a texture inpainter. We showcase here (a) randomly masked texture maps and (b) the inpainted texture maps, with unknown regions rendered as black.}
    \label{fig:inpaint}
    \vspace{-4mm}
\end{figure}

\paragraph{Quantitative comparisons.}
We conduct quantitative comparisons on 400 test objects. Following \cite{siddiqui2022texturify,Yu_2023_ICCV}, we render images from textured meshes and calculate the FID and KID with ground-truth images. As shown in \cref{table:quan}, our method significantly outperforms other methods. Additionally, we tested our model's runtime speed on a single A100 GPU, and our method is notably faster than others, completing evaluations in under 10 seconds without requiring test-time optimization. 

It is worth noting that our method fundamentally differs from others as it is a feed-forward model. Consequently, we can further accelerate our model using techniques such as model compression or diffusion acceleration methods like consistency distillation, which we leave as future work.

\begin{figure}[ht]
    \includegraphics[trim={0cm 0cm 0cm 0cm},clip,width=\columnwidth]{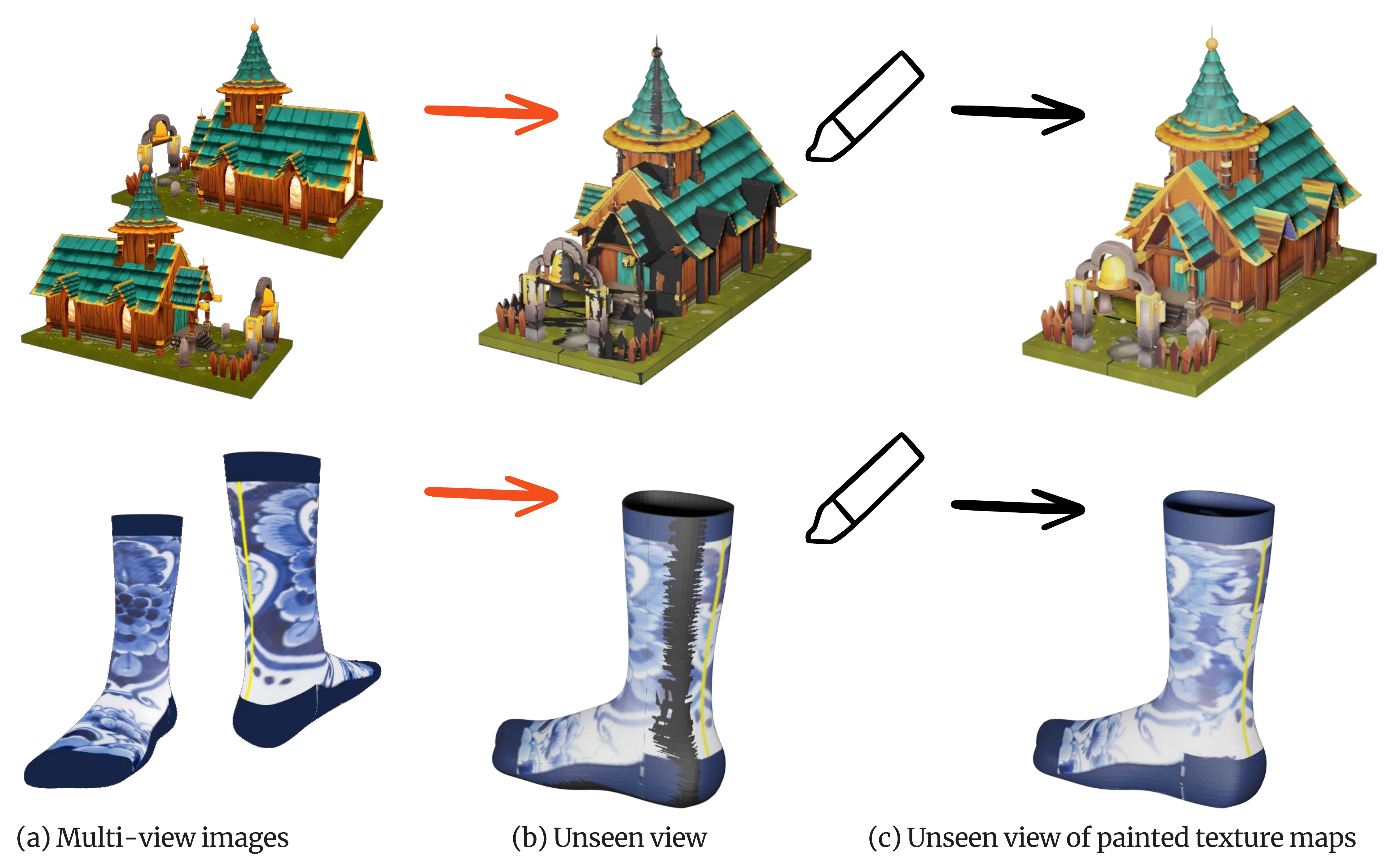}
    \vspace{-6mm}
    \caption{\textbf{Texture completion from sparse views.} With sparse views of objects provided (front and back views as shown in (a)), unprojected textures retain many unseen regions (b). TEXGen effectively fills these unseen regions with harmonious textures (c).}
    \label{fig:sparse_view}
\end{figure}
\subsection{Applications}
Without any fine-tuning, our model serves as a powerful foundation for various applications. Firstly, we showcase the ability of our model to generate mesh textures conditioned solely on text prompts by integrating it with Depth-ControlNet \cite{zhang2023adding}. As shown in \cref{fig:text_condition}, we compose a scene with the generated objects to showcase our results. The scene is vivid and highlights the potential of our model in scene texturing applications. Each object can be customized using individual text prompts for control. 

We also evaluate the text-condition results by two ways. Firstly, we conducted a user study focused on text alignment and texture quality. Participants were asked in each round to choose the best of four texture results based on how well they matched the text description and the quality of the texture. A total of 423 responses were collected, and the final analysis, presented in Table~\ref{table:text_condition_evaluation}, shows which algorithm produced the most preferred results. Besides, we employ the Multi-modal Large Language Model (MLLM) Score~\cite{huang2023t2i} as an objective metric which provides a robust measure of the alignment between the input text prompt and the generated texture, especially under complex conditions that are close to real-world scenarios. As shown in Table~\ref{table:text_condition_evaluation}, both human preference and MLLM score prove that our method outperforms other methods.

\begin{table}[t]
\centering
\caption{\textbf{Quantitative evaluation on text-condition generation.} \textbf{\textit{Preference}} refers to the comprehensive user study evaluating the alignment with the text description and the quality of the texture. For \textbf{\textit{MLLM Score}}, Claude 3.5-sonnet~\cite{anthropic2024claude35}, a state-of-the-art MLLM, is used to calculate the text-to-texture similarity scores. To be consistent with the conclusions in~\cite{huang2023t2icompbench}, we use the same Chain-of-Thought prompts described in the study.}
\vspace{-2mm}
\begin{tabular}{c|cccc}
\toprule
Methods & Paint3D & TEXTure & Text2Tex & Ours \\
\midrule
Preference(\%)($\uparrow$) & 16.5 & 7.1 & 7.1 & \bf{69.3} \\
MLLM Score($\uparrow$) & 64.8 & 69.8 & 64.8 & \bf{74.2}\\
\bottomrule
\end{tabular}
\vspace{-2mm}
\label{table:text_condition_evaluation}
\end{table}

Then, we demonstrate that our model can flexibly paint different sources of partial texture maps. In \cref{fig:inpaint}, we randomly mask sections of a texture map and use our model to fill in the gaps. The results show seamless integration with the existing texture. Additionally, in \cref{fig:sparse_view}, we consider a scenario where the user provides multi-view images. In such cases, there are often many unseen areas. Our model effectively paints these regions, ensuring continuity and coherence.

\subsection{Model Analysis}
\begin{figure}[ht]
    \includegraphics[trim={0cm 0cm 0cm 0cm},clip,width=\columnwidth]{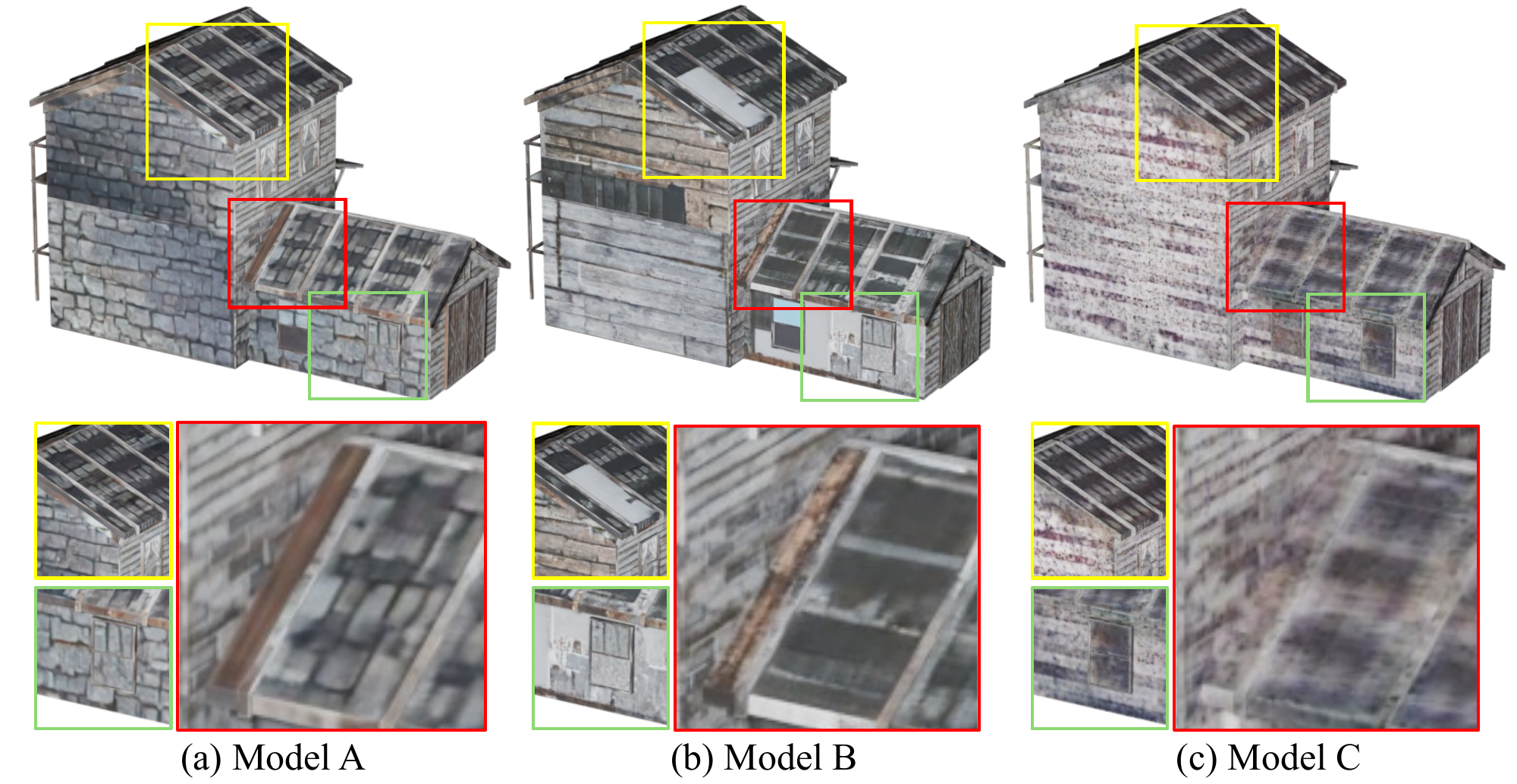}
    \caption{\textbf{Qualitative ablation results on the hybrid design.} Compared to the full model A (a), the model B (b) with only UV blocks can not easily capture overall semantic and 3D consistency while that the model C (c) with only point blocks struggles with producing high-frequency patterns.}
    \label{fig:ablation}
    \vspace{-4mm}
\end{figure}
\paragraph{Hybrid blocks.}
We ablate the hybrid design of our block. For efficiency reasons, we select objects from the house category in our dataset for training and testing in this experiment. We use 10,000 models for training and 100 for testing. We reduce the model size by half to create Model A, which uses the hybrid block. Keeping the parameter count the same as Model A, we remove the point block and replace it with additional UV blocks to create Model B. Similarly, we remove the UV block and replace it with additional point blocks to create Model C, maintaining the same parameter count as Model A. We train all three models on 16 A100 GPUs, with a batch size of 64, for 85K iterations. As seen in \cref{fig:ablation}, compared to Model A, Model B generates textures with inconsistent styles, such as the white patches on the roof in the yellow zoom-in region and the seam artifacts on the wall in the green box. Model C produces relatively consistent textures, but the texture in the red zoom-in region lacks high-frequency details and appears blurred. We also tested FID and KID for the three models. As shown in the \cref{table:abla}, Model A performs the best, confirming the effectiveness of our hybrid block design. 

\paragraph{Classifier-free guidance.}
A key advantage of our model is its use of diffusion-based training instead of a regression loss. This allows us to implement classifier-free guidance~\cite{ho2022classifier} during inference, which is crucial for enhancing texture synthesis quality. We found that the scale of the guidance weight significantly affects the results. While image diffusion models typically use a guidance weight of $\omega = 7.5$ to balance generation quality and condition alignment, our experiments show that a guidance weight of $\omega = 2.0$ provides the optimal balance for our model. As demonstrated in \cref{table:cfg}, we varied the guidance weights and evaluated their performance in terms of FID and KID.

\begin{table}
\centering
\caption{\textbf{Quantitative ablation results on the hybrid design.} Starting from the full model, we build a UV block-only model (B) and a point block-only model (C) by replacing redundant blocks with additional ones, while maintaining the same number of model parameters. FID and KID ($\times 10^{-4}$) are evaluated on multi-view renderings.}
\begin{tabular}{ccc}
\toprule
Models/Metrics & FID($\downarrow$) & KID($\downarrow$) \\
\midrule
Hybrid block (A) & \bf{69.74} &  \bf{17.89} \\
w/o point block (B) & 72.58 & 25.52 \\
w/o UV block (C) & 94.22 & 159.94 \\
\bottomrule
\end{tabular}
\label{table:abla}
\end{table}

\begin{table}
\centering
\caption{\textbf{Ablation results of guidance weights.} We use different CFG weights to evaluate \modelname, and the results show that the weight around 2-3 is optimal.
FID and KID ($\times 10^{-4}$) are evaluated on multi-view renderings.}
\begin{tabular}{c@{\hspace{0.05cm}}c@{\hspace{0.2cm}}c@{\hspace{0.2cm}}c@{\hspace{0.2cm}}c@{\hspace{0.2cm}}c@{\hspace{0.2cm}}c@{\hspace{0.2cm}}c}
\toprule
Metrics/$\omega$ & 1 & 1.5 & 2  & 3 & 4 & 5 & 7.5 \\
\midrule
FID($\downarrow$)  & 35.01 & 34.73 & \bf{34.53} & 35.19 & 35.69 & 36.69 & 39.58 \\
KID($\downarrow$) & 15.06 & 13.00 & 11.94 & \bf{11.71} & 13.03 & 14.53 & 24.45 \\
\bottomrule
\end{tabular}
\label{table:cfg}
\end{table}

\section{Conclusion} 
\label{sec:conc}

In this work, we have presented 
TEXGen, a large generative diffusion model designed for creating high-resolution textures for general 3D objects. 
TEXGen departs from conventional methods that depend on pre-trained 2D diffusion models that necessitate test-time optimization. Instead, our model efficiently synthesizes detailed and coherent textures directly, leveraging a novel hybrid 2D-3D block that adeptly manages both local detail fidelity and global 3D-aware interactions. Capable of generating high-resolution texture maps in a feed-forward manner, 
TEXGen supports a variety of zero-shot applications, including text-guided texture inpainting, sparse-view texture completion, and text-to-texture synthesis. As the first feed-forward model capable of generating textures for general objects, TEXGen sets a new benchmark in the field. We anticipate that our contributions will inspire and catalyze further research and advancements in texture generation and beyond.

\begin{acks}
This work has been supported by Hong Kong Research Grant Council - Early Career Scheme (Grant No. 27209621), General Research Fund Scheme (Grant No. 17202422), Theme-based Research (Grant No. T45-701/22-R) and RGC Matching Fund Scheme (RMGS). Part of the described research work is conducted in the JC STEM Lab of Robotics for Soft Materials funded by The Hong Kong Jockey Club Charities Trust.
\end{acks}

\bibliographystyle{ACM-Reference-Format}
\bibliography{bibs}

\appendix
\section{Appendix}

\subsection{Implementation Details}
We use Objaverse \cite{deitke2023objaverse} as our raw data source, which contains over 800K 3D meshes. However, the texture structure of these meshes is not uniform and requires processing and filtering. For instance, some meshes are divided into parts with multiple texture images, while others have only base color information without texture images. To clean and reorganize the data, we first filter out meshes with poor texture quality. For the remaining meshes, we use xAtlas \cite{JonathanYoung18} to re-unfold the UVs, ensuring they are represented by a single UV atlas. Then, we bake the diffuse color from the original mesh files onto the newly parameterized meshes.
Furthermore, we use Gemini \cite{team2023gemini} to acquire each mesh's caption based on their rendered images. Ultimately, we processed and obtained 120,400 meshes with their corresponding texture images, using 120,000 for training and 400 for evaluation. 
We use five stages to construct our network (i.e., four downsampling and four upsampling stages). For efficiency, we use only UV blocks in the first stage. In the second stage, we use hybrid blocks but replace point attention with sparse convolution. In the remaining three stages, we use our designed hybrid blocks, with the attention layers having 2, 4, and 6 layers, respectively. The number of channels for each stage is 32, 256, 1024, 1024, and 2048, respectively. We use grid sizes of 0.02 and 0.05 in the sparse convolution blocks in the second stage and the attention blocks of the last three stages. The serialized point patch sizes are set to 256, 512, and 1024 in the final three stages.
We train our model using the AdamW optimizer with $\beta_1=0.9, \beta_2=0.999$, and a weight decay of 0.05. The training process takes place on 32 A100 GPUs, with a total batch size of 64, spanning 400K iterations. A cosine scheduler is used to reduce the learning rate from 2e-4 to zero. The code is available at \textcolor{my_href_pink}{\href{https://github.com/CVMI-Lab/TEXGen}{https://github.com/CVMI-Lab/TEXGen}}.

\subsection{More Qualitative Results}
\begin{figure*}[t]
    \includegraphics[trim={0cm 0cm 0cm 0cm},clip,width=\textwidth]{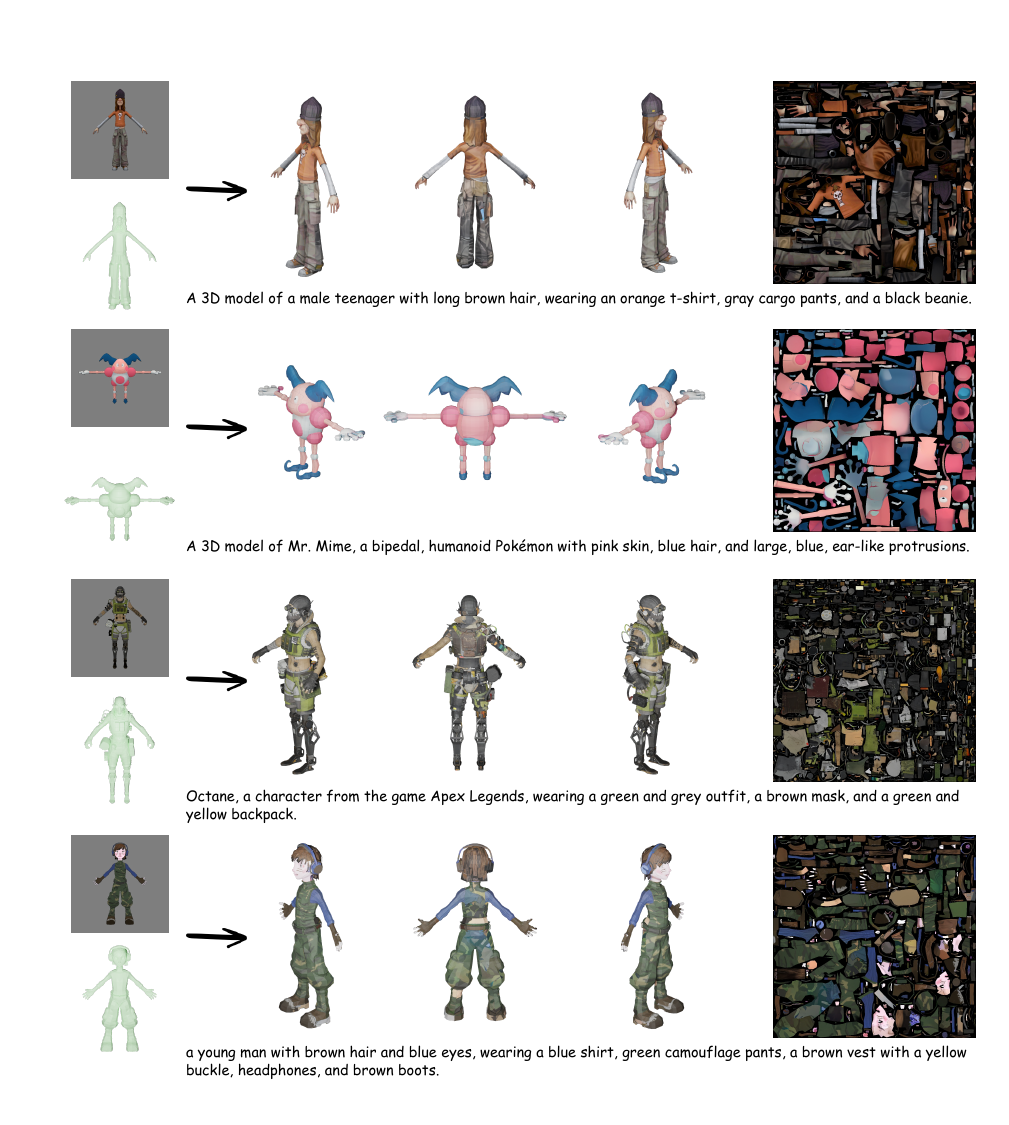}
    \caption{\textbf{Results on 3D avatars.} Our model, trained on 3D data, adeptly avoids the Janus problem.}
    \label{fig:supp_janus}
\end{figure*}

\begin{figure*}[t]
    \includegraphics[trim={0cm 0cm 0cm 0cm},clip,width=\textwidth]{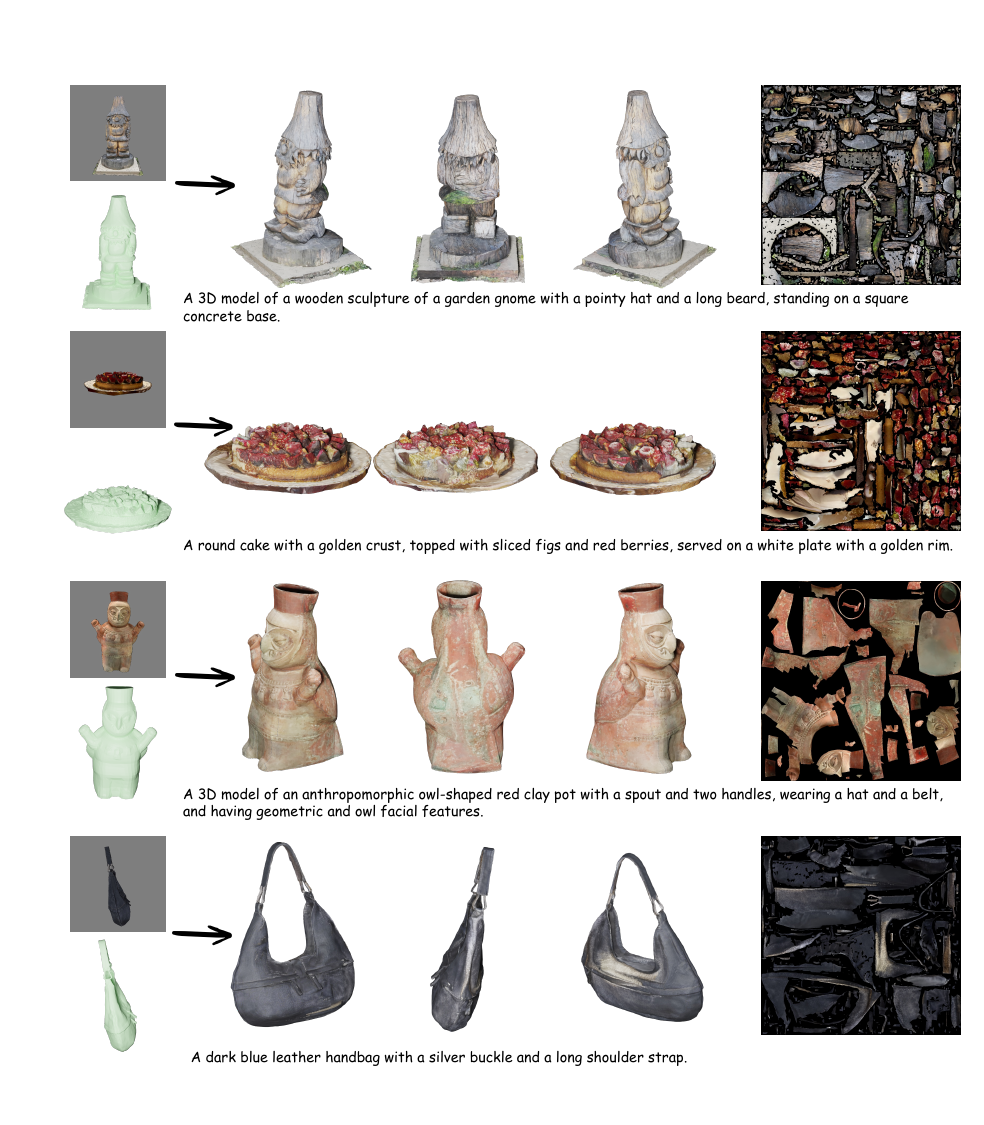}
    \caption{\textbf{Results on real-scan models.} Our method is robust to real-scan models with non-smooth surfaces and fragmented UV maps. }
    \label{fig:supp_real_scan}
\end{figure*}

Firstly, in the video file submitted as supplementary material, we provide a video of the mesh rendering to demonstrate the results of our model. Additionally, our method demonstrates the ability to avoid the Janus problem and is applicable to real-scan models. The Janus problem in texture generation refers to the unintended duplication of features such as eyes and noses on both sides of 3D assets, particularly human faces. This issue typically occurs when methods lack 3D-awareness, often relying on pre-trained image generation models to independently generate textures for different views. Since our model is trained on 3D data and directly produces the full UV map in a forward manner, it effectively prevents this issue, as illustrated in \cref{fig:supp_janus}. Real-scan models often present unique challenges due to their non-smooth surfaces and fragmented, irregular UV maps. Despite these complexities, our method proves robust in handling real-scan models, effectively managing to generate high-quality texture maps that maintain the details and authenticity of the original objects, as demonstrated in \cref{fig:supp_real_scan}.


\subsection{Limitations and Discussions}
Currently, the condition images used during the training of our model are pose-aligned and shape-aligned, which may not meet the needs of users who wish to ``transfer'' textures using arbitrary images. However, we believe our network architecture could potentially handle such scenarios by incorporating dense image information through mechanisms like cross-attention, rather than relying on pixel projection. The primary challenge remains in constructing a suitable dataset for this purpose.
As a future work, we plan to extend our model to generate Physically Based Rendering (PBR) material maps. It can be achieved by collecting and processing data necessary to train the model to generate PBR maps. 

\end{document}